\documentclass{esannV2}

\usepackage[
    paperwidth=21cm,
    paperheight=29.7cm,
    textwidth=12.2cm,
    textheight=19.3cm,
    centering
]{geometry}

\usepackage{graphicx}
\usepackage{amssymb,amsmath,array}
\usepackage{booktabs} 
\usepackage{multirow}
\usepackage{longtable}
\usepackage{hyperref}
\usepackage[capitalize,noabbrev]{cleveref}
\def\ourmethod{AdaCap\,}

\begin{document}
\title{AdaCap: An Adaptive Contrastive Approach for Small-Data Neural Networks}

\author{Bruno Belucci$^1$, Karim Lounici$^2$ and Katia Meziani$^1$
%
\thanks{This project was provided with HPC resources by GENCI at IDRIS/TGCC thanks to the grant 2024-A0181015860.}
\thanks{Submitted to ESANN 2026.}
%
\vspace{.3cm}\\
%
1- Université Paris Dauphine-PSL - CEREMADE \\
Pl. du Maréchal de Lattre de Tassigny, 75016 Paris - France
%
\vspace{.1cm}\\
2- Ecole Polytechnique - CMAP \\
Route de Saclay, 91128 PALAISEAU  - France \\
}

\maketitle


\begin{abstract}
Neural networks struggle on small tabular datasets, where tree-based models remain dominant.  We introduce \textbf{Ada}ptive \textbf{C}ontrastive \textbf{Ap}proach (\ourmethod), a training scheme that combines a permutation-based contrastive loss with a Tikhonov-based closed-form output mapping. Across 85 real-world regression datasets and multiple architectures, \ourmethod yields consistent and statistically significant improvements in the small-sample regime, particularly for residual models. A meta-predictor trained on dataset characteristics (size, skewness, noise) accurately anticipates when \ourmethod is beneficial. 
These results show that \ourmethod acts as a targeted regularization mechanism, strengthening neural networks precisely where they are most fragile. All results and code are publicly available at https://github.com/BrunoBelucci/adacap.
\end{abstract}

\section{Introduction}

Deep neural networks (NNs) achieve state-of-the-art results in vision, language, and large-scale tabular modeling, yet their performance remains unreliable in the \textbf{small-sample regime}, characteristic of many real-world tabular tasks. In contrast, gradient boosting decision trees (GBDTs) remain remarkably robust under limited data, and consistently outperform NNs on curated tabular benchmarks \cite{mcelfreshWhenNeuralNets2023,grinsztajnWhyTreebasedModels2022} and the references cited therein. 
Understanding why NNs collapse in this setting, and how to mitigate this collapse, remains an open problem.

Several approaches have attempted to improve NN robustness on tabular data: architectural modifications inspired by tree ensembles \cite{wydmanskiHyperTabHypernetworkApproach2023}, hyperparameter tuning \cite{kadraWelltunedSimpleNets2021}, regularization strategies \cite{shavittRegularizationLearningNetworks2018}, Bayesian inference \cite{hollmannTabPFNTransformerThat2022}, and data augmentation \cite{brigatoCloseLookDeep2021}. In parallel, contrastive objectives \cite{huComprehensiveSurveyContrastive2024} have emerged as powerful tools for learning structure from limited supervision. Yet, none of these techniques directly address a key challenge of small-data tabular regression: \textbf{NNs must learn a meaningful representation while simultaneously avoiding overfitting}, a trade-off for which standard training pipelines offer little guidance.

We introduce \textbf{\ourmethod}, a simple training scheme designed to strengthen neural networks \textit{precisely where they are most fragile}: in low-data, high-variance regimes. \ourmethod combines two components.
$(i)$ A \textbf{Tikhonov layer} that replaces the standard output layer with a closed-form Tikhonov-regularized solution, providing a stable and data-efficient mapping between the learned representation and the target.
$(ii)$ A \textbf{permutation-based contrastive loss} that increases the fit to the true labels while decreasing the fit obtained when the labels are randomly shuffled, thereby enforcing that the model captures genuine input–target dependencies rather than patterns independent of the inputs.

Across \textbf{85 real-world regression datasets} and diverse architectures 
, \ourmethod produces consistent improvements in small-sample settings. The gains are particularly strong and frequent for \textbf{residual architectures}.
Finally, we show that the effectiveness of \ourmethod is \textbf{predictable}: an XGBoost meta-classifier trained on dataset characteristics (e.g., number of instances, target skewness, noise levels) reaches $\approx 70\%$ accuracy in predicting whether \ourmethod will improve a given model. This indicates that \ourmethod does not act as a universal booster, but rather as a \textbf{targeted and learnable regularizer} whose impact depends on identifiable structural properties of the dataset.

\section{AdaCap Training Scheme}

\ourmethod combines two complementary components:
$(i)$ a \textbf{Tikhonov-based closed-form} output mapping that replaces the standard learned final layer, and
$(ii)$ a \textbf{permutation-based contrastive loss} that contrasts the fit obtained with true versus shuffled labels.
Together, these components provide a strong regularization signal in low-data regimes while preserving compatibility with standard neural architectures.

\textbf{Tikhonov output mapping.}
Let $H \in \mathbb{R}^{n \times d}$ be the final hidden representation produced by the network and let $Y \in \mathbb{R}^n$ be the target vector.
In a standard network, the output layer computes $\widehat{Y} = H W$ for trainable weights $W \in \mathbb{R}^{d}$.
AdaCap replaces these weights by:
$
W(\lambda) = (H^\top H + \lambda I_d)^{-1} H^\top Y$,
$\widehat{Y} = H W(\lambda),
$
where $\lambda \in \mathbb{R}^{+}$ is trainable.
Using the SVD of $H^\top H$, this computation requires only a single matrix factorization and thus remains computationally efficient.


\textbf{Permutation-based contrastive loss.}
To add a contrastive signal, we generate $P$ independent permutations
$\pi_1(Y), \ldots, \pi_P(Y)$
by randomly shuffling the target vector. Each permutation preserves the marginal distribution of $Y$ while destroying any input-output dependency.
For each permuted vector, we reuse the same Tikhonov mapping:
$
W^{(p)}(\lambda) = (H^\top H + \lambda I_d)^{-1} H^\top \pi_p(Y)$,
$\widehat{Y}^{(p)} = H W^{(p)}(\lambda).
$
The Tikhonov solutions all reuse the same SVD of $H^\top H$, avoiding repeated inversions.
The \ourmethod loss contrasts the fit under true labels with the average fit under permuted labels:
$
\mathcal{L}_{\mathrm{AdaCap}}=
\| Y - \widehat{Y} \|-\frac{1}{P}
\sum_{p=1}^P\| \pi_p(Y) - \widehat{Y}^{(p)} \|.
$

This loss is contrastive: it increases when the model fits the true labels better than their shuffled counterparts. Because both fits are computed using the same Tikhonov estimator, the parameter $\lambda$ implicitly scales the contrast by controlling how strongly the estimator responds to structure in the representation $H$.

\textbf{Initialization of the Tikhonov parameter.} 
The performance of \ourmethod is sensitive to the initial value of $\lambda$.
We initialize it by performing a single forward pass over the training set (without updating the network) and evaluating the loss for values of $\lambda$ on a logarithmic grid ranging from $10^{-3}$ to $10^{3}$. 
We pick the value that yields the largest variation in the loss.
This procedure provides a stable starting point and consistently improves subsequent training.

\section{Experimental Results}

\textbf{Datasets.}
We evaluate \ourmethod on 85 real-world regression datasets from the OpenML-CTR23 benchmark\cite{fischerOpenMLCTR23CuratedTabular2023,grinsztajnWhyTreebasedModels2022}. The datasets span a broad range of sizes 
and dimensionalities,
with heterogeneous proportions of numerical and categorical features. A summary of dataset characteristics is provided in \cref{tab:dataset_quartiles1}.
\begin{table}[http]
\begin{footnotesize}
\begin{center}
\caption{Dataset characteristics summary (85 datasets).}
\label{tab:dataset_quartiles1}
\begin{tabular}{lcccccc}
\toprule
Characteristic & Min & Q1 & Median & Q3 & Max & Mean \\
\midrule
Instances ($N$) & 103 & 768 & 8,192 & 21,263 & 5,465,575 & 106,355 \\
Features ($k$) & 3 & 8 & 11 & 21 & 525 & 41 \\
Categorical ($C$) & 0 & 0 & 0 & 3 & 359 & 7 \\
\bottomrule
\end{tabular}
\end{center}
\end{footnotesize}
\end{table}
%
For each dataset, we follow the official train-test splits and use 10-fold cross-validation to report RMSE, MAE, MAPE, and $\text{R}^2$.

\textbf{Effect of the number of permutations.} We first study how the number of permutations $P$ influences performance.  
A simple MLP trained with \ourmethod was evaluated with 
$P \in \{1,2,5,10,20\}$, and for each configuration, we counted the number of wins across all datasets, folds, and metrics.  
As reported in \cref{tab:permutation_win_counts}, the setting $P=10$ achieves the highest overall number of wins.  
We therefore fix $P=10$ for all subsequent experiments.
\begin{table}[http!]
\begin{footnotesize}
\begin{center}
\caption{Win counts and percentages across different permutation levels.}
\label{tab:permutation_win_counts}
\begin{tabular}{rrrrrrr}
\toprule
$P$ & RMSE & MAE & MAPE & $\text{R}^2$ & Total Wins & \% Wins \\
\midrule
1 & 147 & 141 & 164 & 147 & 599 & 17.41 \\
2 & 138 & 126 & 134 & 138 & 536 & 15.58 \\
5 & 151 & 151 & 170 & 151 & 623 & 18.11 \\
10 & \textbf{228} & 215 & \textbf{197} & \textbf{228} & \textbf{868} & \textbf{25.23} \\
20 & 196 & \textbf{227} & 195 & 196 & 814 & 23.66 \\
\bottomrule
\end{tabular}
\end{center}
\end{footnotesize}
\end{table}
%

\textbf{Models.} We consider MLP and ResNet architectures \cite{heDeepResidualLearning2015,kadraWelltunedSimpleNets2021}
with four layers and a 256-dimensional hidden size, using 256-dimensional embeddings for categorical variables and ReLU activations.
Variants include: (i) {Deeper} (8 layers), (ii) {GLU} (Gated Linear Units \cite{dauphinLanguageModelingGated2017}), and (iii) {GReLUOneCycleLR} (Generalized ReLU with One-Cycle learning rate scheduling \cite{smithSuperConvergenceVeryFast2018}). 
We also evaluate a vanilla Transformer \cite{vaswaniAttentionAllYou2017}. 
All architectures are trained both with and without \ourmethod.

\begin{figure}[http]
\centering
\includegraphics[width=\columnwidth]{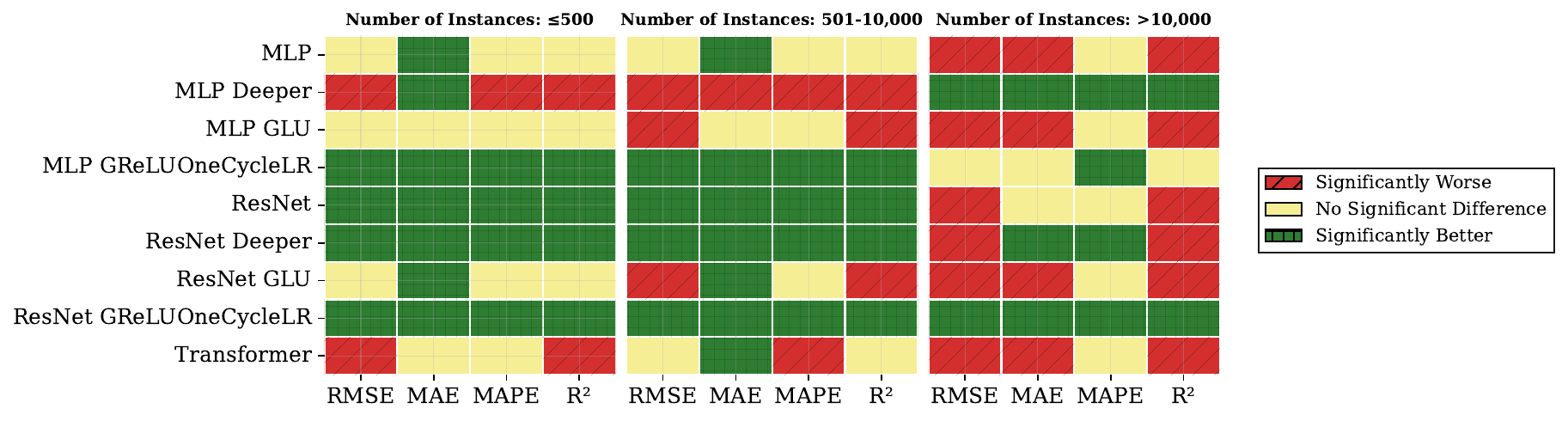}
\caption{Wilcoxon outcomes across dataset-size regimes.}
\label{fig:wilcoxon}
\end{figure}

\textbf{Impact Across Architectures and Dataset Sizes.}
We next evaluate the effect of \ourmethod across architectures and dataset sizes. \cref{fig:wilcoxon}  summarizes the Wilcoxon signed-rank outcomes comparing each model with and without \ourmethod across RMSE, MAE, MAPE, and $\text{R}^2$.

Across dataset sizes, the effect of \ourmethod is strong but not uniform. In the \textbf{small-data regime} ($N < 500$), \ourmethod consistently improves performance for several architectures: MLP GReLUOneCycleLR, ResNet, ResNet Deeper, and ResNet GReLUOneCycleLR are all significantly better across every metric. The plain MLP and ResNet GLU also benefit in this regime, although their gains are mostly limited to MAE. This confirms that \ourmethod is particularly effective when data are scarce and the representation needs stronger regularization.

In the \textbf{intermediate regime} ($500 < N \le 10{,}000$), the effect becomes more heterogeneous. Residual models still improve in many cases, but some variants (e.g., ResNet GLU) begin to show mixed outcomes across metrics. Certain MLP variants shift from improvements to degradations, and Transformers exhibit a metric-dependent behavior, ranging from neutral to mildly positive or negative.

For \textbf{large datasets} ($N > 10{,}000$), \ourmethod is no longer uniformly beneficial. 
MLP Deeper becomes consistently better across all metrics, while the plain MLP is often neutral or worse. Some ResNet variants lose their improvements, except ResNet GReLUOneCycleLR, which remains strongly positive. Transformers are mostly unaffected or degraded in this regime.

Overall, these results indicate that \ourmethod is not a universal booster: \textbf{its strongest gains appear in small-data settings and in specific architecture–size combinations}. 
This structured behavior motivates the following analysis on predicting when \ourmethod is most effective.

\section{Predicting When AdaCap Helps}

A key question is to predict \emph{a priori} when \ourmethod will improve a given architecture on a given dataset. 
For each (dataset, architecture) pair, we build a feature vector of dataset characteristics: number of samples and features; categorical-variables counts and cardinalities; summary statistics of numerical variables (skewness, kurtosis, correlations); numerical-features outlier ratios; intrinsic dimensionality; PCA variance ratios; target skewness and kurtosis; and a noise estimate from a random forest regressor. Target label indicates whether \ourmethod improves the metric relative to the corresponding base model.

\begin{table}[h]
\begin{footnotesize}
\begin{center}
\caption{Top 10 features ranked by SHAP importance.}
\label{tab:top_10_features_sum_of_ranks}
\begin{tabular}{lr}
\toprule
Feature & Sum of Ranks \\
\midrule
Number of Instances & 224 \\
Target Skewness & 538 \\
Noise Estimation (RF) & 588 \\
Min Skewness of Numerical Features & 613 \\
Max Absolute Correlation of Numerical Features & 630 \\
Target Kurtosis & 634 \\
Min Absolute Correlation of Numerical Features & 637 \\
Min Kurtosis of Numerical Features & 644 \\
Outlier Ratio of Numerical Features & 668 \\
Mean Skewness of Numerical Features & 674 \\
\bottomrule
\end{tabular}
\end{center}
\end{footnotesize}
\end{table}
\begin{figure}[http!]
\centering
\includegraphics[height=0.35\textheight]{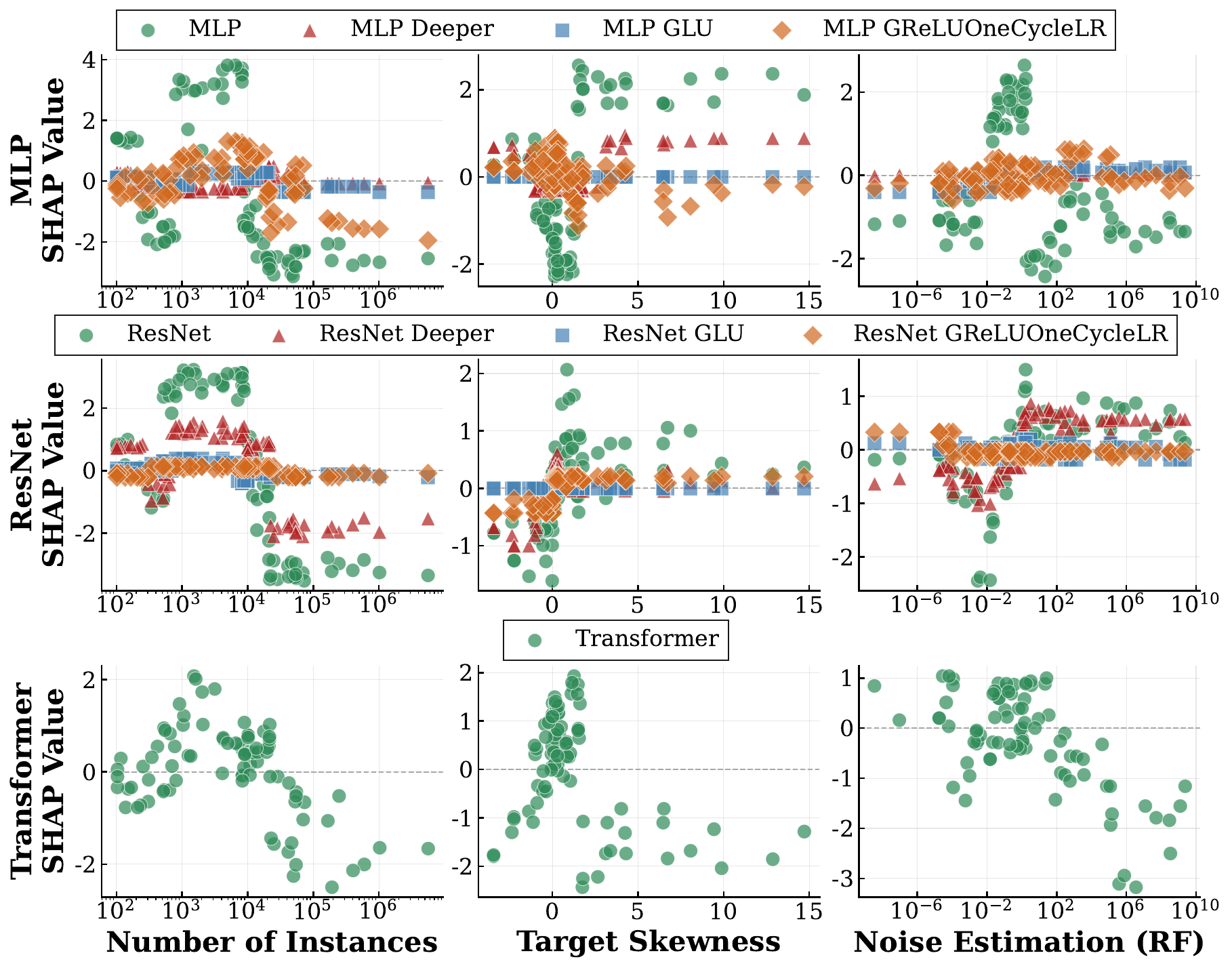}
\caption{SHAP values of the top dataset characteristics for several architectures.}
\label{fig:shap-values}
\end{figure}

To assess predictability, we train a gradient-boosted tree classifier across all 85 datasets for each model. 
The classifiers achieve a mean accuracy of 70\% (std.\ 11\%) on held-out runs with a 20\% test split, showing that improvements from \ourmethod are indeed predictable from dataset structure. Restricting the classifier to only the top three features, according to the sum of their rank using the mean absolute SHAP \cite{lundbergUnifiedApproachInterpreting2017} values (\cref{tab:top_10_features_sum_of_ranks}), still yields an accuracy of 67\% (std.\ 12\%), indicating that a small subset of dataset characteristics is sufficient for reliable predictions.

\cref{fig:shap-values} presents SHAP scatter plots for several architectures. 
The dominant predictor across all models is \textbf{dataset size}: smaller datasets make improvements from \ourmethod substantially more likely. 
Transformers benefit more from datasets with symmetric target distributions, whereas MLPs and ResNets benefit more strongly from positively skewed targets. 
Noise level also interacts with architecture: \ourmethod tends to help ResNets in noisy datasets, MLPs in low-noise regimes, and Transformers in low to medium noise settings.

\section{Conclusion}

We presented \ourmethod, a simple combination of a Tikhonov closed-form output and a permutation-based contrastive loss that strengthens neural networks in small-data tabular regression. 
Experiments on 85 datasets show substantial gains for residual architectures in low-sample regimes, with more heterogeneous effects for larger datasets. 
A meta-classifier accurately predicts when \ourmethod helps, revealing dataset size, target skewness, and noise level as dominant factors. 
\ourmethod thus provides a targeted and interpretable improvement where neural networks are most brittle.




\begin{footnotesize}



\bibliographystyle{unsrt}
\bibliography{adacap_esann.bib}

\end{footnotesize}


\end{document}